# Detecting floodwater on roadways from image data with hand-crafted features and deep transfer learning*

Cem Sazara[1], Mecit Cetin[1] and Khan M. Iftekharuddin[2]

*Abstract*— Detecting roadway segments inundated due to floodwater has important applications for vehicle routing and traffic management decisions. This paper proposes a set of algorithms to automatically detect floodwater that may be present in an image captured by mobile phones or other types of optical cameras. For this purpose, image classification and flood area segmentation methods are developed. For the classification task, we used Local Binary Patterns (LBP), Histogram of Oriented Gradients (HOG) and pre-trained deep neural network (VGG-16) as feature extractors and trained logistic regression, k-nearest neighbors, and decision tree classifiers on the extracted features. Pre-trained VGG-16 network with logistic regression classifier outperformed all other methods. For the flood area segmentation task, we investigated superpixel based methods and Fully Convolutional Neural Network (FCN). Similar to the classification task, we trained logistic regression and k-nearest neighbors classifiers on the superpixel areas and compared that with an end-to-end trained FCN. Conditional Random Fields (CRF) method was applied after both segmentation methods to post-process coarse segmentation results. FCN offered the highest scores in all metrics; it was followed by superpixel-based logistic regression and then superpixel-based KNN.

## I. INTRODUCTION

Flood-prone communities, e.g., coastal cities, experience frequent flooding due to storm surge, heavy rain, and sea level rise. According to one study, by 2050, recurrent flooding will be common for many US coastal cities and will likely occur thirty or more days per year due to sea-level rise [1]. Information about roadway inundation is critical during the flooding events because delivery of goods and services and movement of emergency vehicles depend on a working transportation infrastructure. This paper provides a method for monitoring roadway inundation based on crowdsourced images with the help of machine learning methods. Machine learning has become popular recently, mainly due to the progress made in deep learning methods.

Earlier literature in flood detection has mainly focused on using remote sensing devices such as satellite and aircraft [2, 3]. Although remote sensing methods provide high-level flood information, they lack local details. Greater local details can give more information about the severity, extent, and depth of floodwater on specific road segments. Video and surveillance-based methods have also been studied to detect flood or water level changes [4, 5]. Water segmentation in images and videos is a related problem to flood detection. Water area detection was studied for unmanned vehicle navigation [6, 7, 8]. Spatiotemporal information and probabilistic models were also studied to detect water in videos [9-10]. Ground-level flood detection and classification have been an active research field recently. Geetha et al. proposed a method to find the floodwater extend and approximate depth of it using thresholding techniques [13]. Lopez-Fuentes et al. proposed a multi-modal deep learning method for flood image classification using textual information and images from social media [14]. Dry-flood image comparison-based methods were also studied [15, 16]. These models used hand-selected parameters and compared flood-dry image pairs for each flood location.

In this paper, our main contributions are:

- *Development of an image classifier system to classify images as flood images or dry images*. Different from the work of Lopez-Fuentes et al. [14], we didn't incorporate metadata and only used image information. In addition, we compared deep learning and classical machine learning methods for feature extraction.

- *Fully automated flood area segmentation using superpixel based and deep learning methods on a diverse crowdsourced dataset.* Our models don't require hand-selected thresholds as was done in some previous studies [13, 15, 16]. The methods are tested on images collected in different environments such as urban, suburban, and natural settings.

- *Generation of a dataset of 253 hand-labeled flood images coming from different locations with different characteristics*. This dataset is used both in image classification and flood area segmentation tasks. The dataset is made available by the authors to support further research in this field °.

For the classification task, we used the following methods for feature extraction: Local Binary Patterns (LBP), Histogram of Oriented Gradients (HOG), and pre-trained VGG-16 deep learning network. We used logistic regression, k-nearest

*This study was supported by the Mid-Atlantic Transportation Sustainability Center (MATS UTC), Region 3 UTC funded by USDOT

[1]Cem Sazara and Mecit Cetin are with Transportation Research Institute, Old Dominion University, Norfolk, VA 23529, email: csaza001@odu.edu, mcetin@odu.edu

[2]Khan M. Iftekharuddin is with Electrical and Computer Engineering department, Old Dominion University, Norfolk, VA 23529, email: kiftekha@odu.edu

°DataLink: https://www.dropbox.com/sh/grxeep1k9a0yziq/AAByrZYB-jGQoTvb0Yp22fJFa

neighbors, and decision trees as classification methods on the extracted features. For this task, we used 253 flood images and 238 images without floodwater. For the flood area segmentation task, we used superpixel-based handcrafted features and Fully Convolutional Neural Network structure [17]. Some background information about these methods is provided in Section II. In Section III, we show sample images and briefly discuss the image data. Implementation section explains our model training procedures, and in Section IV, we summarize the performance and accuracy of the models.

## II. BACKGROUND

### A. Feature Extractors

Feature extractors are used to construct feature vectors in machine learning problems. For this purpose, we used Local Binary Patterns (LBP), Histogram of Oriented Gradients (HOG), and pre-trained neural network.

**Local Binary Patterns (LBP):** LBP method was introduced by Ojala et al 1994 [18]. It is used as a visual descriptor for texture classification tasks. In this method, image is divided into cells of the same size, e.g., 16x16 pixels. Each pixel in the cell is compared with its neighbors following a clockwise or anti-clockwise circle. If the pixel value is greater than a neighboring pixel value, we place 1 (and 0 otherwise). This gives a 4-digit binary number for 4 connected neighbors (top, down, right and left) and 8-digit binary number for 8 connected neighbors (top, top-left, top-right, down, down-left, down-right, right, left). These binary numbers will be converted to decimals and their values are kept in a histogram for each cell in the image. When we concatenate all histograms coming from the cells, we have a single feature vector for the image.

**Histogram of Oriented Gradients (HOG):** Histogram of Oriented Gradients (HOG) is a feature extraction method that models appearance or shape of objects in an image [19]. In this method, image gradients in X and Y directions are calculated. Gradient histograms are calculated and normalized for separate cells. When all histograms coming from the cells are concatenated, a single feature vector for the image is constructed.

**Pre-trained deep neural network (VGG-Net):** VGG-net was proposed by Simonyan and Zisserman in 2015 for image classification task [20]. VGGNet achieved state-of-the art results in ImageNet Large Scale Visual Recognition Challenge (ILSVRC) [21]. VGGNet introduced a simple structure with the same convolution and max-pooling layers applied many times. It consists of multiple 3x3 convolutions followed by max-pooling layers. There are 16- and 19- layer versions of this network. In this paper, we used VGG-16 as a feature extraction method and got output of the last max-pooling layer in the network. Using a pre-trained network on another problem is called "Transfer Learning". This helps transfer knowledge already acquired (from large datasets) to another problem (with smaller dataset) and thereby reducing the training time. Use of pre-trained CNNs for image classification has been a common way to achieve good results.

Some roadway and transportation related CNN applications are ranging from vehicle classification [22-25, 35], road damage detection [26] to transportation speed prediction [27].

### B. Classifiers

Classifiers are machine learning models that are trained with the extracted features described in Section A. We used three common supervised classification methods, i.e., logistic regression, k-nearest neighbors, and decision trees, to predict whether or not a given image contains a scene with floodwater.

**Logistic Regression:** Logistic regression is a classification method where input variables are multiplied by coefficients and mapped to discrete classes using logistic sigmoid functions [28]. Logistic sigmoid functions provide probability values that can map to discrete values for classification. Sigmoid function $\rho(t)$ is given in Eq. 1.

$$\rho(t) = \frac{1}{1+e^{-t}} \quad (1)$$

Usually 'L1' or 'L2' regularization is applied as penalty against model complexity. This regularization reduces the weight coefficients and helps achieve better generalization performance.

**K-Nearest Neighbors (KNN):** In K-Nearest Neighbors classifier, a data point is classified based on the majority of the "K" nearest training data points [28]. "K" is a positive number and needs be determined beforehand (hyperparameter). If K is selected too small, the model can include noise in the data and may not perform well. If K is too large, the model doesn't reflect the general properties of the training data. KNN is considered an instance-based learning method.

**Decision Trees (DTs):** Decision tree (DT) is a flowchart structure that creates rules similar to if-else statements. The tree represents the decisions [29]. Decision tree classification logic is easy to interpret for humans as we can trace the path followed in the tree. DTs grow into branches looking for purer subsets at each split (Figure 1).

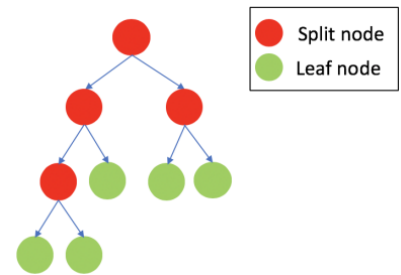

*Figure 1. Decision tree nodes*

DTs usually use cost functions that measure purity level after the split decisions. Some common cost functions are gini, entropy, and classification error.

### C. Semantic Segmentation Networks

The purpose of semantic segmentation is to assign every pixel within the image to a class of interest. In this paper, there are two classes: floodwater and everything else. We studied superpixel and Fully Convolutional Neural Networks (FCNs) based semantic segmentation methods.

**Superpixel based Semantic Segmentation with Hand-crafted Features:** Superpixels are groups of pixels with similar intensity values. Superpixel segmentation divides an image into non-overlapping areas or regions. The main advantages of using this method are reducing the input feature size for subsequent classification algorithms and calculating features over related regions. One disadvantage of this method is that initial segmentation errors will be propagated through the rest of the process. In this paper, we used Simple Linear Iterative Clustering (SLIC) superpixel method [30] for its low computational cost. SLIC uses k-means clustering to efficiently create superpixels. Feature extraction methods are applied to each superpixel region and classifiers are trained to assign the correct class label to each region. More information about the model structure is given in the implementation and results sections.

**Fully Convolutional Neural Networks (FCNs):** FCNs are commonly used for semantic segmentation tasks. They were introduced by Long et. al in 2015 [17]. In classification, conventionally, images go through convolution and pooling layers and then fully connected layers. The output is the class of the whole image (Figure 2). The width and height at each layer get smaller (downsampled).

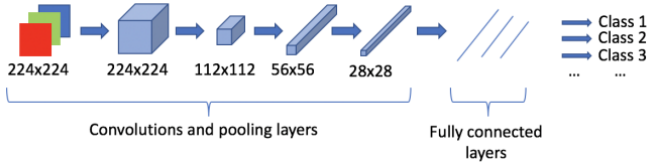

*Figure 2. Conventional image classification network*

A conventional classification network can be converted to FCN by cutting the network before fully connected layers and adding upsampling layers. Different level of information from previous layers can be used with skip connections. Figure 3 below shows the general structure of this network.

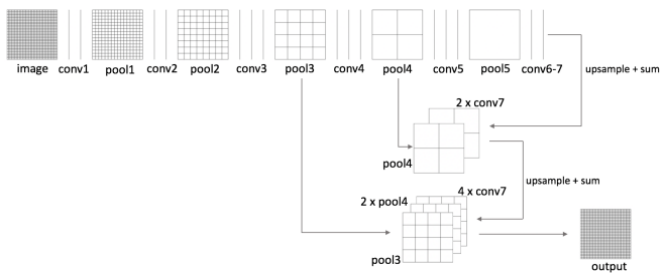

*Figure 3. FCN structure*

After each upsampling layer, image size (width and height) is increased. Upsampled layers are then merged and a single 2D matrix with size equal to the original image size is achieved. In order to reduce training time and the need for a large image database, we can use pre-trained networks for the convolution part of the network. Possible options include AlexNet [31], VGG-Net [20], and GoogLeNet [32] which can be used as the convolutional layers or feature extractors [17]. In this paper, we used VGG-Net.

**Conditional Random Fields (CRFs):** CRFs are used to smooth semantic segmentation results. CRFs are applied after both superpixel based methods and semantic segmentation network in our implementation. Let G be a graph over x and y where x is a vector formed by random variables $x_1, x_2, x_3, ..., x_V$ where V is the number of pixels in the image and Y is a label random variable that can take values from $\{y_1, y_2, …, y_n\}$ where n is the number of classes [33]. x values denote our local image features such as color or texture. Every x has a corresponding y value which denotes its label. CRF models the probability P(y/x) with a Hidden Markov Field where the conditional probability is only dependent on the current position and its adjacent pixels (Markov property) [36]. G = (V, E) is the graph with vertices V and edges E. This relationship is shown in Figure 4.

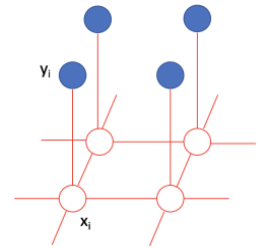

*Figure 4. CRF graph structure*

The CRF energy function is given by Eq. 2. We try to minimize this energy function.

$$E(X) = \sum_{i \in V} \psi(x_i) + \sum_{i,j \in V} \psi(x_i, x_j) \qquad (2)$$

where the first term $\psi(x_i)$ is called the unary term, the second term $\psi(x_i, x_j)$ is the pairwise term and xi and xj are adjacent pixel values. Unary term is assigned by the pixel class probability and the pairwise term accounts for smoothness in adjacent pixels and assigns similar labels to pixels with similar properties [34]. The labels that result in the smallest energy value E(X) are used.

### III. DATASET

Our dataset consists of 253 flood images and 238 images without flood. All images have the same size of 385x512. Flood images were hand-labeled so that pixels corresponding to flood areas have value of one and the rest of the pixels are zero. The flood image dataset contains different scenes from urban, suburban and natural settings and it will be useful for further flood detection-segmentation research. Some sample images are shown below in Figure 5. Flooded areas are marked with yellow color.

### IV. IMPLEMENTATION

#### A. Classification

For the classification task, we used 253 flood images and 238 non-flood images. Dataset is shuffled and split into training

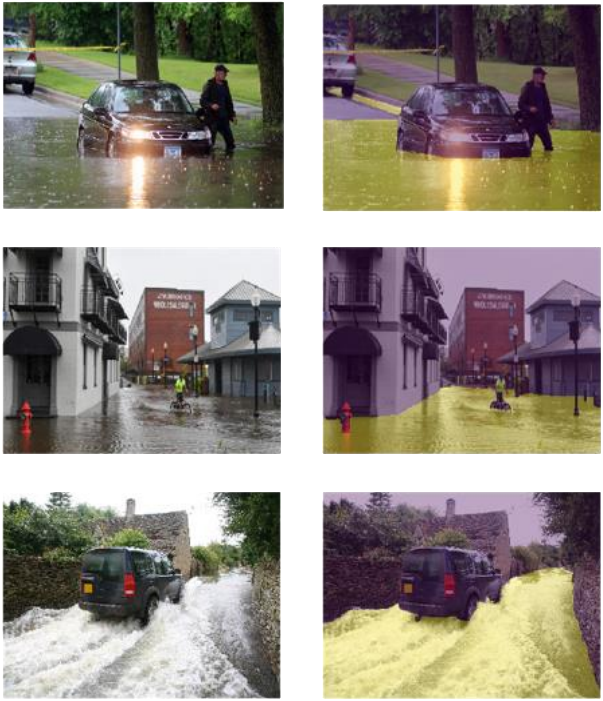

*Figure 5. Sample flood images and their pixel labels (floodwater in yellow)*

(80%) and test (20%) subsets. We used three feature extractors: LBP, HOG and VGG-16. After feature extraction step, we trained three classifiers: Logistic regression, k-nearest neighbors and decision trees for each of the feature extraction. This gives a total of 9 classifiers. We used grid search with 5-fold cross validation for hyperparameter optimization. Average inference time for each classification method is primarily affected by the feature extraction method. For a Macbook Pro computer with Intel I5 processor and 8GB memory, LBP takes 0.2 seconds, HOG takes 10.3 seconds and VGG takes 1.25 seconds to process a single image. Images are resized to 224x224 for the classification task. Results are summarized in Table 2 and discussed in the next section.

*B. Semantic Segmentation*

For this task, we used the 253 flood images. Similar to the classification task, we split the dataset into training (80%) and test (20%) subsets.

**Superpixel and Hand-crafted Features:** We implemented superpixel-based methods and hand-crafted feature vectors. Feature vectors include mean x and y positions, R intensity, G intensity, B intensity, and local binary patterns (LBP) histogram for each superpixel area. We trained logistic regression and k-nearest neighbors classifiers on these features. SLIC with 250 segments were used as the superpixel segmentation method and we used grid search for hyper-parameter optimization of the classifiers. Average inference time for a single image on a Macbook Pro computer with Intel I5 processor and 8GB memory is 2.12 seconds.

**Fully Convolutional Neural Network (FCN):** For the initial convolution part of the FCN, we used a pre-trained VGG-16. We fine-tuned and trained the network with the training subset and tested on the test subset. Stochastic gradient descent with learning rate 0.01 and Nesterov momentum were used in the optimization process. We used Keras deep learning library with Python 3.6. The model was trained for 30 epochs with batch size 12. Average inference time for a single image is 2.27 seconds with a MacBook Pro computer with Intel I5 processor and 8GB memory. Results are summarized in Table 3 and discussed next.

## V. RESULTS

In this section, we summarize our test results. We used precision, recall and F1-scores as performance measures which are derived from the values in the confusion matrix. Confusion matrix is defined in TABLE 1.

*Table 1. Confusion matrix*

|  |  | Prediction | |
|---|---|---|---|
|  |  | $\hat{y} = 0$ | $\hat{y} = 1$ |
| True label | y=0 | True Negative | False Positive |
|  | y=1 | False Negative | True Positive |

Our performance measures are given in Eq. 3-5 below.

$$Precision = \frac{\sum True\ Positive}{\sum True\ Positive + \sum False\ Positive} \quad (3)$$

$$Recall = \frac{\sum True\ Positive}{\sum True\ Positive + \sum False\ Negative} \quad (4)$$

$$F1 = 2\ x\ \frac{Precision\ x\ Recall}{Precision + Recall} \quad (5)$$

We summarize classification and semantic segmentation results in this section.

*A. Classification*

Three different feature extraction methods (LBP, HOG and VGG-16) and three classifiers (logistic regression, k-nearest neighbors and decision tree) were used for the binary classification of images (i.e., whether an image contains a scene with floodwater). VGG-16 deep neural network feature extraction achieves the highest scores in all measures: 0.94 precision, 0.97 recall, and 0.95 F1-score.

*B. Semantic Segmentation*

In this section, we compared superpixel hand-crafted measures and fully convolutional neural network. FCN produced the highest scores in all metrics. Similarly, logistic regression achieved higher scores than KNN in all metrics.
Some sample FCN results are shown in Figure 6. Detected flood areas are marked with yellow color. Looking at our test results, we can realize that the model sometimes misclassifies pixels that are reflections of objects on the water. We will further work on improving the model for reflection cases.

*Table 2. Classification Results*

| Feature extraction | Classifier | Precision | Recall | F1-score |
|---|---|---|---|---|
| LBP | Logistic Regression | 0.76 | 0.72 | 0.74 |
| LBP | KNN | 0.63 | 0.76 | 0.69 |
| LBP | Decision Tree | 0.61 | 0.68 | 0.64 |
| HOG | Logistic Regression | 0.70 | 0.82 | 0.76 |
| HOG | KNN | 0.56 | 0.88 | 0.69 |
| HOG | Decision Tree | 0.71 | 0.60 | 0.65 |
| VGG-16 | Logistic Regression | **0.94** | **0.97** | **0.95** |
| VGG-16 | KNN | 0.67 | 0.89 | 0.77 |
| VGG-16 | Decision Tree | 0.80 | 0.77 | 0.79 |

*Table 3. Semantic Segmentation Results*

| Feature extraction | Classifier | Precision | Recall | F1-score |
|---|---|---|---|---|
| Superpixel and hand-crafted features | Logistic Regression | 0.89 | 0.84 | 0.86 |
| Superpixel and hand-crafted features | KNN | 0.83 | 0.82 | 0.82 |
| Fully Convolutional Neural Network (FCN) | | **0.92** | **0.90** | **0.91** |

## VI. CONCLUSION

In this paper, we studied flood image classification and flood area segmentation problems and provided a new pixel labeled dataset. We introduced novel flood image classification methods. VGG-16 deep learning-based feature extraction with logistic regression resulted in high scores on all performance measures (precision, recall and F1-score). For flood area segmentation, we compared superpixel-based methods with FCN. In this task, the proposed methods resulted in closer results to each other than the methods for the classification task. FCN has further improvement potential with more labeled data. As a future work, more complex semantic segmentation networks will be tested against the superpixel based methods and FCN. In addition, the authors will investigate extracting more information from the floodwater such as flood severity level and water depth estimation and work on improving the model for water reflection cases.

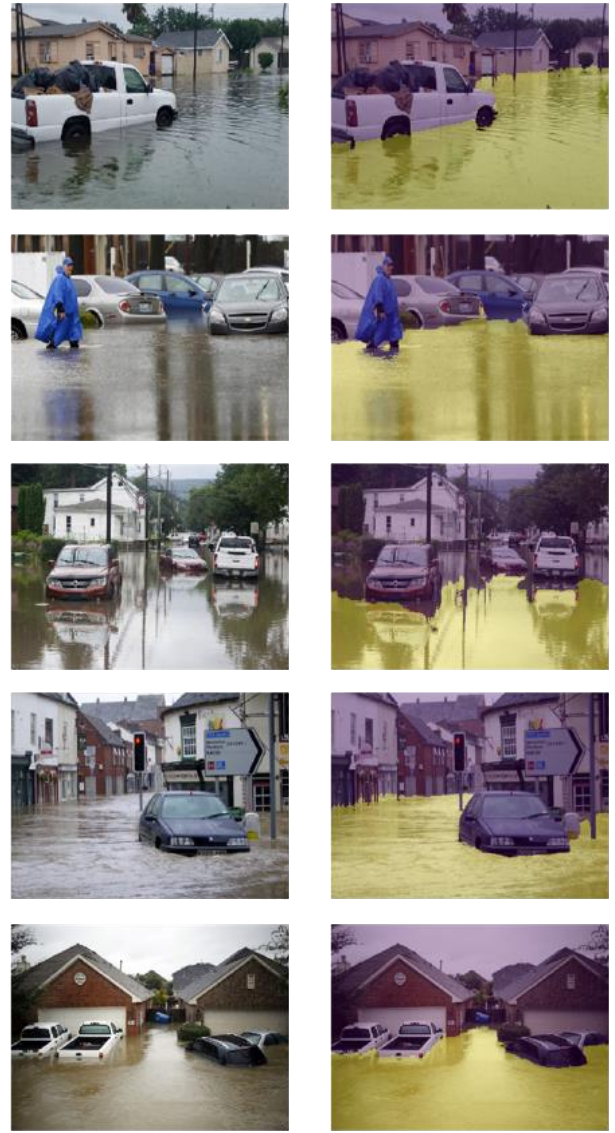

*Figure 6. Some FCN flood segmentation results*

REFERENCES

[1] W. V. Sweet and J. Park, "From the extreme to the mean: Acceleration and tipping points of coastal inundation from sea level rise," *Earths Future*, vol. 2, no. 12, pp. 579–600, Dec. 2014.
[2] V. Klemas, "Remote Sensing of Floods and Flood-Prone Areas: An Overview," *J. Coast. Res.*, pp. 1005–1013, Dec. 2014.
[3] N. M. Robertson and T. Chan, "Aerial image segmentation for flood risk analysis," in *2009 16th IEEE International Conference on Image Processing (ICIP)*, 2009, pp. 597–600.
[4] C. L. Lai, J. C. Yang, and Y. H. Chen, "A Real Time Video Processing Based Surveillance System for Early Fire and Flood Detection," in *2007 IEEE Instrumentation Measurement Technology Conference IMTC 2007*, 2007, pp. 1–6.
[5] S.-W. Lo, J.-H. Wu, F.-P. Lin, and C.-H. Hsu, "Visual Sensing for Urban Flood Monitoring," *Sensors*, vol. 15, no. 8, pp. 20006–20029, 2015.
[6] M. Iqbal, O. Morel, and F. Meriaudeau, "A survey on water hazard detection," 2009, pp. 33–39.

[7] A. Rankin and L. Matthies, "Daytime water detection based on color variation," in *2010 IEEE/RSJ International Conference on Intelligent Robots and Systems*, 2010, pp. 215–221.

[8] A. L. Rankin, L. H. Matthies, and A. Huertas, "Daytime water detection by fusing multiple cues for autonomous off-road navigation," in *Transformational Science and Technology for the Current and Future Force*, vol. Volume 42, 0 vols., WORLD SCIENTIFIC, 2006, pp. 177–184.

[9] P. V. K. Borges, J. Mayer, and E. Izquierdo, "A probabilistic model for flood detection in video sequences," in *2008 15th IEEE International Conference on Image Processing*, 2008, pp. 13–16.

[10] P. Mettes, R. T. Tan, and R. Veltkamp, "On the segmentation and classification of water in videos," in *2014 International Conference on Computer Vision Theory and Applications (VISAPP)*, 2014, vol. 1, pp. 283–292.

[11] P. Mettes, R. T. Tan, and R. C. Veltkamp, "Water detection through spatio-temporal invariant descriptors," *Comput. Vis. Image Underst.*, vol. 154, pp. 182–191, Jan. 2017.

[12] P. Santana, R. Mendonça, and J. Barata, "Water detection with segmentation guided dynamic texture recognition," in *2012 IEEE International Conference on Robotics and Biomimetics (ROBIO)*, 2012, pp. 1836–1841.

[13] M. Geetha, M. Manoj, A. S. Sarika, M. Mohan, and S. N. Rao, "Detection and estimation of the extent of flood from crowd sourced images," in *2017 International Conference on Communication and Signal Processing (ICCSP)*, 2017, pp. 0603–0608.

[14] L. Lopez-Fuentes, J. van de Weijer, M. Bolaños, and H. Skinnemoen, "Multi-modal deep learning approach for flood detection," in *Working Notes Proceedings of the MediaEval 2017 Workshop CEUR-WS*, 2017, pp. 1–3.

[15] M. A. Witherow, M. I. Elbakary, K. M. Iftekharuddin, and M. Cetin, "Analysis of Crowdsourced Images for Flooding Detection," in *VipIMAGE 2017*, 2018, pp. 140–149.

[16] M. A. Witherow, C. Sazara, I. M. Winter-Arboleda, M. I. Elbakary, M. Cetin, and K. M. Iftekharuddin, "Floodwater detection on roadways from crowdsourced images," *Comput. Methods Biomech. Biomed. Eng. Imaging Vis.*, vol. 0, no. 0, pp. 1–12, Jun. 2018.

[17] J. Long, E. Shelhamer, and T. Darrell, "Fully Convolutional Networks for Semantic Segmentation," *ArXiv14114038 Cs*, Nov. 2014.

[18] T. Ojala, M. Pietikainen, and D. Harwood, "Performance evaluation of texture measures with classification based on Kullback discrimination of distributions," in *Proceedings of 12th International Conference on Pattern Recognition*, 1994, vol. 1, pp. 582–585 vol.1.

[19] N. Dalal and B. Triggs, "Histograms of oriented gradients for human detection," in *2005 IEEE Computer Society Conference on Computer Vision and Pattern Recognition (CVPR'05)*, 2005, vol. 1, pp. 886–893 vol. 1.

[20] K. Simonyan and A. Zisserman, "Very Deep Convolutional Networks for Large-Scale Image Recognition," *ArXiv14091556 Cs*, Sep. 2014.

[21] "ImageNet Large Scale Visual Recognition Challenge (ILSVRC)." [Online]. Available: http://www.image-net.org/challenges/LSVRC/.

[22] R. V. Nezafat, B. Salahshour, and M. Cetin, "Classification of truck body types using a deep transfer learning approach," in *2018 21st International Conference on Intelligent Transportation Systems (ITSC)*, 2018, pp. 3144–3149.

[23] R. Vatani Nezafat, O. Sahin, and M. Cetin, "A Deep Transfer Learning Approach for Classification of Truck Body Types Based on Side-Fire Lidar Data," presented at the Transportation Research Board 98th Annual MeetingTransportation Research Board, 2019.

[24] E. Gundogdu, E. S. Parıldı, B. Solmaz, V. Yücesoy, and A. Koç, "Deep learning-based fine-grained car make/model classification for visual surveillance," in *Counterterrorism, Crime Fighting, Forensics, and Surveillance Technologies*, 2017, vol. 10441.

[25] J. T. Lee and Y. Chung, "Deep Learning-Based Vehicle Classification Using an Ensemble of Local Expert and Global Networks," in *2017 IEEE Conference on Computer Vision and Pattern Recognition Workshops (CVPRW)*, 2017, pp. 920–925.

[26] H. Maeda, Y. Sekimoto, T. Seto, T. Kashiyama, and H. Omata, "Road Damage Detection and Classification Using Deep Neural Networks with Smartphone Images," *Comput.-Aided Civ. Infrastruct. Eng.*, vol. 33, no. 12, pp. 1127–1141, 2018.

[27] X. Ma, Z. Dai, Z. He, J. Ma, Y. Wang, and Y. Wang, "Learning Traffic as Images: A Deep Convolutional Neural Network for Large-Scale Transportation Network Speed Prediction," *Sensors*, vol. 17, no. 4, Apr. 2017.

[28] S. Shalev-Shwartz and S. Ben-David, *Understanding Machine Learning: From Theory to Algorithms*, 1 edition. New York, NY, USA: Cambridge University Press, 2014.

[29] J. R. Quinlan, "Simplifying decision trees," *Int. J. Man-Mach. Stud.*, vol. 27, no. 3, pp. 221–234, Sep. 1987.

[30] R. Achanta, A. Shaji, K. Smith, A. Lucchi, P. Fua, and S. Süsstrunk, "SLIC Superpixels Compared to State-of-the-Art Superpixel Methods," *IEEE Trans. Pattern Anal. Mach. Intell.*, vol. 34, no. 11, pp. 2274–2282, Nov. 2012.

[31] A. Krizhevsky, I. Sutskever, and G. E. Hinton, "ImageNet Classification with Deep Convolutional Neural Networks," in *Advances in Neural Information Processing Systems 25*, F. Pereira, C. J. C. Burges, L. Bottou, and K. Q. Weinberger, Eds. Curran Associates, Inc., 2012, pp. 1097–1105.

[32] C. Szegedy *et al.*, "Going Deeper with Convolutions," *ArXiv14094842 Cs*, Sep. 2014.

[33] H. Zhou, Jun Zhang, Jun Lei, Shuohao Li, and Dan Tu, "Image semantic segmentation based on FCN-CRF model," in *2016 International Conference on Image, Vision and Computing (ICIVC)*, 2016, pp. 9–14.

[34] A. Torralba, K. Murphy, W. T. Freeman, and M.A. Rubin, "Context-based vision system for place and object recognition," in *Proceedings Ninth IEEE International Conference on Computer Vision*, 2003, pp. 273–280 vol.1.

[35] R.V. Nezafat, O. Sahin and M. Cetin, "Transfer Learning Using Deep Neural Networks for Classification of Truck Body Types Based on Side-Fire Lidar Data," *Journal of Big Data Analytics*, vol. 1, no. 1, pp. 71-82, 2019

[36] T. Liu, X. Huang and J. Ma, "Conditional Random Fields for Image Labeling," *Mathematical Problems in Engineering*, 2016, http://dx.doi.org/10.1155/2016/3846125